\documentclass[11pt,a4paper]{article}
\usepackage[hyperref]{emnlp2018}
\usepackage{times}
\usepackage{latexsym}
\usepackage{amsmath,array,graphicx}
\usepackage{amssymb}
\usepackage{tabularx}
\usepackage{boldline}
\usepackage{multirow}
\usepackage{lscape}
\usepackage{pbox}
\usepackage{tabulary}

\renewcommand{\vec}{\boldsymbol}

\usepackage{url}

\aclfinalcopy 
\def\aclpaperid{265} 

\title{Hierarchical Neural Networks for Sequential Sentence Classification in Medical Scientific Abstracts}

\author{Di Jin \\
  MIT \\
  {\tt jindi15@mit.edu} \\\And
  Peter Szolovits \\
  MIT \\
  {\tt psz@mit.edu} \\}

\date{}

\begin{document}
\maketitle
\begin{abstract}

Prevalent models based on artificial neural network (ANN) for sentence classification often classify sentences in isolation without considering the context in which sentences appear. This hampers the traditional sentence classification approaches to the problem of sequential sentence classification, where structured prediction is needed for better overall classification performance. In this work, we present a hierarchical sequential labeling network to make use of the contextual information within surrounding sentences to help classify the current sentence. Our model outperforms the state-of-the-art results by 2\%-3\% on two benchmarking datasets for sequential sentence classification in medical scientific abstracts. 

\end{abstract}

\section{Introduction}

Since 1665, over 50 million scholarly research articles have been published \cite{jinha2010article}, with approximately 2.5 million new scientific papers coming out each year \cite{ware2015stm}. While this enormous corpus provides us with the ability to conclusively accept or reject hypotheses and yields insight into promising research directions, it is getting harder and harder to extract useful information from the literature in an efficient and timely manner due to its sheer amount. Therefore, an automatic and intelligent tool to help users locate the information of interest quickly and comprehensively is highly desired. 

When searching for relevant literature for a certain field, investigators first check the abstracts of scientific papers to see whether they match the criterion of interest. This process can be expedited if the abstracts are structured; that is, if the rhetorical structural elements of scientific abstracts such as \textit{purpose, methods, results, and conclusions} \cite{american1979american} are explicitly stated. However, even  today, a significant portion of scientific abstracts is still unstructured, which causes great difficulty in information retrieval. In this paper, we develop a machine-learning based approach to automatically categorize sentences in scientific abstracts into rhetorical sections so that the desired information can be efficiently retrieved.

In a scientific abstract, each sentence can be assigned to a rhetorical structural element sequentially. This rhetorical structure profiling process can be formulated as a sequential sentence classification task, as the element assignment of any single sentence is greatly associated with the assignments of the surrounding sentences. This is in contrast to the general sentence classification problem, where each sentence is classified individually and no contextual information can be used. Previous state-of-the-art methods relied on Conditional Random Fields (CRFs) to take into account the inter-dependence between subsequent labels, which improved joint sentence classification performance by considering the label sequence information. In this work, we add a bi-directional long short-term memory (bi-LSTM) layer over the representations of individual sentences so that it can encode the contextual content and semantics from preceding and succeeding sentences for better categorical inference of the current one.

In this work, we present a hierarchical neural network model for the sequential sentence classification task, which we call a hierarchical sequential labeling network (HSLN). Our model first uses a RNN or CNN layer to individually encode the sentence representation from the sequence of word embeddings, then uses another bi-LSTM layer to take as input the individual sentence representation and output the contextualized sentence representation, subsequently uses a single-hidden-layer feed-forward network to transform the sentence representation to the probability vector, and finally optimizes the predicted label sequence jointly via a CRF layer. 
%
%
%
We evaluate our model on two benchmarking datasets, PubMed RCT \cite{dernoncourt2017pubmed} and NICTA-PIBOSO \cite{kim2011automatic}, which were both generated from the PubMed database\footnote{https://www.ncbi.nlm.nih.gov/pubmed/}. 
%
%
Our key contributions are summarized as follows:
\begin{enumerate}
\item Based on the previous best performing architecture for sequential sentence classification \cite{dernoncourt2016neural}, we add one more layer to extract contextual information from surrounding sentences for more accurate prediction of the current one. Together with the CRF algorithm, this allows us to make use of not only the preceding labels' information but also the content and semantics of adjacent sentences to infer the label of the target sentence. 
%
%
%
%
\item We remove the need for a character-based word embedding component without sacrificing  performance. For individual sentence encoding, we propose the use of a CNN module as an alternative to RNN for small datasets, suffering less from over-fitting as evidenced by our experiments. 
%
%
%
%
%
%
%
%
Moreover,  we incorporate attention-based pooling in both RNN and CNN models to further improve the performance. 
%
%
%
%
\item We adopt dropout with expectation-linear regularization instead of the standard one to reduce the performance gap between training and test phases.
\item We obtain state-of-the-art results on two datasets for sequential sentence classification in medical abstracts, outperforming the previous best models by at least 2\% in terms of F1 scores.
\end{enumerate}

\section{Related Work}

Previous systems for sequential sentence classification concentrate on the rhetorical structure analysis of biomedical abstracts. They are mainly based on naive Bayes \cite{ruch2007using}, support vector machine (SVM) \cite{mcknight2003categorization,yamamoto2005sentence,liu2013abstract}, Hidden Markov Model (HMM) \cite{lin2006generative}, and CRF \cite{kim2011automatic,hassanzadeh2014identifying,hirohata2008identifying,chung2009sentence}. All these methods heavily rely on numerous carefully hand-engineered features such as lexical (bag-of-words (BOW)), semantic (hypernyms, synonyms), structural (part of speech (POS) tags, lemmas, orthographic shapes, headings), statistical (statistical distributions of token types) and sequential (sentence position, surrounding features, predicted labels) features. 
%
%

In contrast, current emerging artificial neural network (ANN) based models have removed the need for manually selected features; instead, features are self-learned from the token and/or character embeddings. These deep learning models have revolutionized the natural language processing (NLP) field with state-of-the-art results achieved in various tasks, including the most relevant text classification task \cite{kim2014convolutional,zhang2016dependency,conneau2017very,lai2015recurrent,joulin2016bag,ma2015dependency}. Most of these models are built upon deep CNNs or RNNs as well as  combinations of them, where CNN is good at extracting local n-gram features while RNN is suitable for sequence modeling.
%
%
%
%
%
%

The above-mentioned works for short-text classification do not consider any context of sentence semantics in the models, 
%
%
making them under-perform in the sequential sentence classification scenario, where surrounding sentences can play a big role in inferring the label of the current sentence. Recent works that apply deep neural networks to the sequential sentence classification problem include the system proposed by Lee et al. \cite{lee2016sequential}, where the preceding utterances were used to help classify the current utterance in a dialog into the corresponding dialogue act. Most recent work from Dernoncourt et al. \cite{dernoncourt2016neural} used a CRF layer to optimize the predicted label sequence, where the preceding labels have influence on determining the current label. This model outperformed the state-of-the-art results on two datasets PubMed RCT and NICTA-PIBOSO for sentence classification in medical abstracts.

\section{Proposed Model}

\paragraph{Notation} We denote scalars in italic lowercase (e.g., $k$), vectors in bold italic lowercase (e.g., $\vec{s}$) and matrices in italic uppercase (e.g., $W$). Colon notations $x_{i:j}$ and $\vec{s}_{i:j}$ are used to denote the sequence of scalars $(x_i, x_{i+1}, ..., x_{j})$ and vectors $(\vec{s}_i, \vec{s}_{i+1},..., \vec{s}_j)$.
%
%

Our model is composed of four components: the word embedding layer, the sentence encoding layer, the context enriching layer, and the label sequence optimization layer. In the following sections they will be discussed in detail.

\subsection{Word Embedding Layer}

Given a sentence $\vec w = \begin{bmatrix} w_1 & w_2 & \cdots & w_N\end{bmatrix}$ comprising $N$ words, this layer maps each word to a real-valued vector as its lexical-semantic representation. Word representations are encoded by the column vector in the embedding matrix $W^{word}\in \mathbb{R}^{d^w\times|V|}$, where $d^w$ is the dimension of the word vector and $V$ is the vocabulary of the dataset. Each column $W_i^{word}\in \mathbb{R}^{d^w}$ is the word embedding vector for the $i^{th}$ word in the vocabulary.  The word embeddings $W^{word}$ can be pre-trained on large unlabeled datasets using unsupervised algorithms such as word2vec \cite{mikolov2013distributed}, GloVe \cite{pennington2014glove} and fastText \cite{bojanowski2016enriching}.
%
%
%
%

\subsection{Sentence Encoding Layer}

This layer takes as input the embedding vector of each token in a sentence from the word embedding layer and produces a vector $\textbf{s}$ to encode this sentence. The sequence of embedding vectors is first processed by a bi-directional RNN (bi-RNN) or CNN layer, similar to the ones used in the text classification before \cite{kim2014convolutional,lee2016sequential,liu2016recurrent}. This layer outputs a sequence of hidden states $\vec{h}_{1:N}$ ($\vec{h}\in \mathbb{R}^{d^{hs}}$) for a sentence of $N$ words with each hidden state corresponding to a word. To form the final representation vector $\vec{s}$ of this sentence, attention-based pooling is used, which can be described using the following equations:
%
%
%
%
%
%

\begin{equation}
A = \text{softmax}(U_s\tanh(W_sH+\vec{b}_s)),
\end{equation}
\begin{equation}
S=AH^T,
\end{equation}
where $H=\begin{bmatrix} h_1 & h_2 & \cdots & h_N\end{bmatrix}\in \mathbb{R}^{d^{hs}\times N}$, $W_s\in \mathbb{R}^{d^a \times d^{hs}}$ is the transformation matrix for soft alignment, $\vec{b}_s\in \mathbb{R}^{d^a}$ is the bias vector, $U_s\in \mathbb{R}^{r \times d^{a}}$ is the token level context matrix used to measure the relevance or importance of each token with respect to the whole sentence, softmax is performed along the second dimension of its input matrix, and $A\in \mathbb{R}^{r \times N}$ is the attention matrix.

Here each row of $U_s$ is a context vector $\vec{u}_s\in \mathbb{R}^{d^{a}}$ and it is expected to reflect an aspect or component of the semantics of a sentence. To represent the overall semantics of the sentence, we use multiple context vectors to focus on different parts of this sentence. 
%
%

Finally, the sentence encoding vector $\vec{s}\in \mathbb{R}^{rd^{hs}}$ is obtained by reshaping the matrix $S$ into a vector.

\begin{figure}[h!]
\begin{center}
\includegraphics[width=0.5\textwidth]{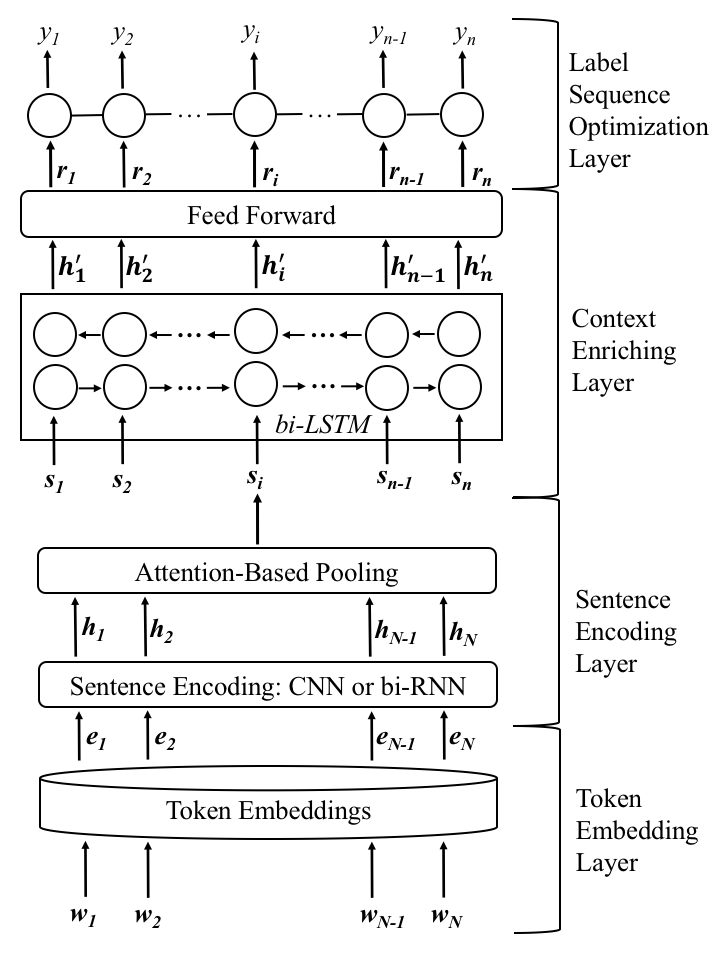}
\caption{Model architecture. $\vec{w}$: original word; $\vec{e}$: word embedding vector; $\vec{h}$: sentence-level hidden state output by the bi-RNN or CNN layer; $\vec{s}$: sentence representation vector; $\vec{h'}$: abstract-level hidden state output by the bi-LSTM layer; $\vec{r}$: sentence label probability vector; $y$: predicted sentence label.}
\label{figure:model}
\end{center}
\end{figure}

\subsection{Context Enriching Layer}

This layer takes as input the sequence of individual sentence encoding vectors in a given abstract of $n$ sentences obtained from the last sentence encoding layer, with each vector corresponding to a sentence. It outputs a new sequence of contextualized sentence encoding vectors, which are enriched with the contextual information from surrounding sentences. Specifically, the sequence of individual sentence encoding vectors is input into a bi-LSTM layer, which produces a sequence of hidden state vectors $\vec{h'}_{1:n}$ ($\vec{h'}\in \mathbb{R}^{d^{hd}}$) with each corresponding to a sentence. Each of these vectors is subsequently input to a feed-forward neural network with only one hidden layer to get the corresponding probability vector $\vec{r}\in \mathbb{R}^{l}$, which represents the probability that this sentence belongs to each label, where $l$ is the number of labels.
%
%
%
%
%
%
%
%
%
%

\subsection{Label Sequence Optimization Layer}

Within the abstract, the sequence of sentence categories implicitly follows some patterns. For example, the category \textit{Results} is always followed by \textit{Conclusion}, and the category \textit{Methods} is certainly after the \textit{Background}. Making use of such patterns can boost the classification performance via the CRF algorithm \cite{Lample2016NeuralAF}. Given the sequence of probability vectors $\vec{r}_{1:n}$ from the last context enriching layer for an abstract of $n$ sentences, this layer outputs a sequence of labels $y_{1:n}$, where $y_i$ represents the predicted label assigned to the $i^{th}$ sentence.
%
%
%
%
%

In the CRF algorithm, in order to model dependencies between subsequent labels, we incorporate a matrix $T$ that contains the transition probabilities between two subsequent labels; we define $T[i, j]$ as the probability that a token with label $i$ is followed by a token with the label $j$. The score of a label sequence $y_{1:n}$ is defined as the sum of the probabilities of individual labels and the transition probabilities:
\begin{equation}
s(y_{1:n})=\sum_{i=1}^n \vec{r}_i(y_i)+\sum_{i=2}^n T[y_{i-1}, y_i].
\end{equation}

The score in the above equation can be transformed into the probability of a certain label sequence by taking a softmax operation over all possible label sequences:

\begin{equation}
p(y_{1:n})=\frac{e^{s(y_{1:n})}}{\sum_{\hat{y}_{1:n}\in Y}e^{s(\hat{y}_{1:n})}},
\end{equation}
where $Y$ denotes the set of all possible label sequences. During the training phase, the objective is to maximize the probability of the gold label sequence. In the testing phase, given an input sequence, the corresponding sequence of predicted labels is chosen as the one that maximizes the score, computed via the Viterbi algorithm \cite{forney1973viterbi}.
%
%

\section{Experiments}

\subsection{Datasets}

We evaluate our model on two sources of benchmarking datasets on medical scientific abstracts, where each sentence of the abstract is annotated with one label that is associated with the rhetorical structure. Table \ref{table:data-statistics} summarizes the statistics of the two datasets.
%
%
%
%

\paragraph{NICTA-PIBOSO}

This dataset\footnote{This dataset can be found online at https://www.kaggle.com/c/alta-nicta-challenge2} was shared from the ALTA 2012 Shared Task \cite{amini2012overview}, the goal of which is to build automatic sentence classifiers that can map the sentences from biomedical abstracts into a set of pre-defined categories for Evidence-Based Medicine (EBM).

\paragraph{PubMed RCT}

This new dataset was curated by \cite{dernoncourt2017pubmed}\footnote{This dataset can be downloaded from https://github.com/Franck-Dernoncourt/pubmed-rct} and is currently the largest dataset for sequential sentence classification. It is based on the PubMed database of biomedical literature and each sentence of each abstract is labeled with its role in the abstract using one of the following classes: \textit{background, objective, method, result, and conclusion}. Table \ref{table:abstract-example} presents an example abstract comprising structured sentences with their annotated labels. 
%
%
%
%

\begin{table*}[h!]
\centering
\begin{tabular}{|l|c|c|c|c|c|}
\hline
\textbf{Dataset} & \textbf{$|C|$} & \textbf{$|V|$} & Train & Validation & Test \\ \hline
NICTA-PIBOSO            & 6            & 17k          & 720 (7.7k)       & 80 (0.9k)           & 200 (2.2k)      \\ \hline
PubMed 20k       & 5            & 68k          & 15k (180k)     & 2.5k (30k)          & 2.5k (30k)    \\ \hline
PubMed 200k      & 5            & 331k         & 190k (2.2M)    & 2.5k (29k)          & 2.5k (29k)    \\ \hline
\end{tabular}
\caption{Datasets statistics. $|C|$ denotes the number of labels, $|V|$ represents the vocabulary size. For the train, validation, and test sets, we indicate the number of abstracts followed by the number of sentences in parentheses.}
%
%
\label{table:data-statistics}
\end{table*}

\begin{table*}[h!]
\centering
\begin{tabular}{ |>{\raggedright\arraybackslash} m{2.7cm} |>{\raggedright\arraybackslash} m{12.3cm}|}
\hline
\textbf{Category} & \textbf{Sentences}                                                                                                                                                                                              \\ \hline
BACKGROUND                            & Emotional eating is associated with overeating and the development of obesity. {[}...{]}                                    \\ \hline
OBJECTIVES                         & The aim of this study was to test if attention bias for food moderates the effect of self-reported emotional eating during sad mood (vs neutral mood) on actual food intake. [...]                                        \\ \hline
METHODS                        & Participants (N = 85) were randomly assigned to one of the two experimental mood induction conditions (sad/neutral). [...]            \\ \hline
RESULTS                   & [...] Yet, attention maintenance on food cues was significantly related to increased intake specifically in the neutral condition, but not in the sad mood condition.                                                                   \\ \hline
CONCLUSIONS               & The current findings show that self-reported emotional eating (based on the DEBQ) might not validly predict who overeats when sad, at least not in a laboratory setting with healthy women. [...]                                \\ \hline
\end{tabular}
\caption{A typical abstract example with structured sentences and their corresponding annotated labels. The PMID of this abstract is 24854809. }
\label{table:abstract-example}
\end{table*}

\subsection{Training Settings}

For both datasets, test performance is assessed on the training epoch with best validation performance and F1 scores (weighted average by support (the number of true instances for each label)) are reported as the results.
%
%
%
%

The token embeddings were pre-trained on a large corpus combining Wikipedia, PubMed, and PMC texts \cite{moen2013distributional} using the word2vec tool\footnote{The word vectors can be downloaded at http://bio.nlplab.org/} (denoted as ``Word2vec-wiki+P.M.''). They are fixed during the training phase to avoid over-fitting. We also tried other types of word embeddings, such as the word2vec embeddings pre-trained on the Google News dataset\footnote{https://code.google.com/archive/p/word2vec/} (denoted as ``Word2vec-News''), word2vec embeddings pre-trained on the Wikipedia corpus\footnote{https://github.com/jind11/word2vec-on-wikipedia} (denoted as ``Word2vec-wiki''), GloVe embeddings pre-trained on the corpus of Wikipedia 2014 + Gigaword 5\footnote{http://nlp.stanford.edu/data/glove.6B.zip} (denoted as ``Glove-wiki''), fastText embeddings pre-trained on Wikipedia\footnote{https://github.com/facebookresearch/fastText/blob/master/ pretrained-vectors.md} (denoted as ``FastText-wiki''), and fastText embeddings initialized with the standard GloVe Common Crawl embeddings and then fine-tuned on PubMed abstracts plus MIMIC-III notes (denoted as ``FastText-P.M.+MIMIC''). The comparison results are summarized in the next section.
%
%

The model is trained using the Adam optimization method \cite{kingma2014adam}. The learning rate is initially set as 0.003 and decayed by 0.9 after each epoch. For regularization, dropout \cite{srivastava2014dropout} is applied to each layer. For the version of dropout used in practice (e.g., the dropout function implemented in the TensorFlow and Pytorch libraries), the model ensemble generated by dropout in the training phase is approximated by a single model with scaled weights in the inference phase, resulting in a gap between training and inference. To reduce this gap, we adopted the dropout with expectation-linear regularization introduced by \citet{ma2016dropout} to explicitly control the inference gap and thus improve the generalization performance.
%
%
%
%
%
%

Hyperparameters were optimized via grid search based on the validation set and the best configuration is shown in Table \ref{table:hyperparameters}. The window sizes of the CNN encoder in the sentence encoding layer are 2, 3, 4 and 5. The RNN encoder in the sentence encoding layer is set as LSTM for the PubMed datasets and gated recurrent unit (GRU) for the NICTA-PIBOSO dataset. Code for this work is available online\footnote{https://github.com/jind11/HSLN-Joint-Sentence-Classification}.
%
%


\begin{table}[h!]
\centering
\begin{tabular}{lrrrrrr}
\hlineB{2.5}
\multirow{2}{*}{Parameter} & \multicolumn{2}{c}{PubMed}     & \multicolumn{2}{c}{NICTA}                         \\
                                 & \multicolumn{1}{c}{RNN} & \multicolumn{1}{c}{CNN}  & \multicolumn{1}{c}{RNN} & \multicolumn{1}{c}{CNN} \\ \hline
$d^{hs}$                 & 200  & -  & 200 & -         \\
$d^{hd}$                 & 200  & 200  & 200 & 300        \\
$d^a$                    & 200  & 100  & 250 & 75        \\
$d^c$                    & -   & 200  & - & 150        \\
$r$                      & 15  & 1  & 5 & 4      \\
$\beta$                  & 0.01 & 0.001  & 0.01  & 0.01      \\
$dr$                     & 0.5  & 0.5 & 0.6 & 0.6\\\hlineB{2.5}
\end{tabular}
\caption{Hyperparameter settings. $d^{hs}$: hidden size of the sentence-level RNN layer (single direction); $d^{hd}$: hidden size of the abstract-level bi-LSTM layer (single direction); $d^a$: dimension of the context vector $\vec{u}_s$; $r$: number of context vectors; $\beta$: coefficient of the dropout regularization added to the total loss; $dr$: dropout.}
\label{table:hyperparameters}
\end{table}

\section{Results and Discussion}

Table \ref{table:comparison} compares our model against the best performing models in the literature \cite{dernoncourt2016neural,liu2013abstract}. There are two variants of our model in terms of different implementations of the sentence encoding layer: the model that uses bi-RNN to encode the sentence is called HSLN-RNN; while the model that uses the CNN module is named HSLN-CNN. We have evaluated both model variants on all datasets. And as evidenced by Table \ref{table:comparison}, our best model can improve the F1 scores by 2\%-3\% in absolute number compared with the previous best published results for all datasets. For the PubMed 20k and 200k datasets, our HSLN-RNN model achieves better results; however, for the NICTA dataset, the HSLN-CNN model performs better. This makes sense because the CNN sentence encoder has fewer parameters to be optimized, thus the HSLN-CNN model is less likely to over-fit in a smaller dataset such as NICTA. With sufficient data, however, the increased capacity of the HSLN-RNN model offers performance benefits.  To be noted, this performance gap between RNN and CNN sentence encoder gets larger as the dataset size increases from 20k to 200k for the PubMed dataset.

\begin{table}[h!]
\centering
\begin{tabularx}{0.5\textwidth}{>{\hsize=0.8\hsize}X
                              >{\hsize=0.15\hsize\centering\arraybackslash}X
                              >{\hsize=0.15\hsize\centering\arraybackslash}X
                              >{\hsize=0.15\hsize\centering\arraybackslash}X}
\hlineB{2.5}
\multicolumn{1}{l}{\multirow{2}{*}{Model}} & \multicolumn{2}{c}{PubMed} & {\multirow{2}{*}{NICTA}} \\
\multicolumn{1}{l}{}        & 20k          & 200k        &         \\ \hline
\textit{Best Published}    &            &             &       \\
\pbox{10cm}{Marco Lui \\ \cite{lui2012feature}}        & -          & -           & 82.0  \\
\pbox{10cm}{bi-ANN \\ \cite{dernoncourt2016neural}}     & 90.0       & 91.6        & 82.7  \\ \hline
\textit{Our Models} &            &             &       \\
HSLN-CNN            &  92.2      &   92.8          &  \textbf{84.7}     \\
HSLN-RNN            &  \textbf{92.6} & \textbf{93.9}   &  84.3    \\ \hlineB{2.5}
\end{tabularx}
\caption{Comparison of F1 scores (weighted average by support (the number of true instances for each label)) between our model and the best published methods. The presented results of our model are evaluated on the test set of the run with the highest F1 score on the validation set.}
\label{table:comparison}
\end{table}

Table \ref{table:ablation-study} presents the ablation analysis of our model (on the PubMed 20k dataset), where we remove one component at a time and quantify the performance drop (reported on F1 scores). As can be seen from Table \ref{table:ablation-study}, our HSLN-CNN model uniformly suffers a little more from the component removal than the HSLN-RNN model, indicating that the HSLN-RNN model is more robust. When the context enriching layer is removed, both models experience the most significant performance drop and can only be on par with the previous state-of-the-art results, strongly demonstrating that this proposed component is the key to the performance improvement of our model. Furthermore, even without the label sequence optimization layer, our model still significantly outperforms the best published methods that are empowered by this layer, indicating that the context enriching layer we propose can help optimize the label sequence by considering the context information from the surrounding sentences. Last but not the least, the dropout regularization and attention-based pooling components we add to our system can help further improve the model in a limited extent. 
%
%
%
%
%
%
%
%
%

\begin{table}[h!]
\centering
\begin{tabular}{lrr}
\hlineB{2.5}
{Model} & HSLN-RNN & HSLN-CNN \\ \hline
Full Model     &  \textbf{92.6}          &  \textbf{92.2}              \\
$-$ context      &  90.0      &  89.0              \\
$-$ seq. opt.    &  92.3      &  91.8              \\
$-$ dropout reg. &  92.4      &  91.9              \\
$-$ attention    &  92.4      &  91.7              \\\hlineB{2.5}
\end{tabular}
\caption{Ablation analysis. F1 scores are reported. ``$-$ context'' is our model without the context enriching layer. ``$-$ seq. opt.'' is our model without the label sequence optimization layer. ``$-$ dropout reg.' is our model using the standard dropout strategy without the expectation-linearization regularization. ``$-$ attention'' refers to the model without attention-based pooling, i.e., in the sentence encoding layer, the final hidden state is used for the HSLN-RNN model while max-pooling is used for the HSLN-CNN model.}
\label{table:ablation-study}
\end{table}

Table \ref{table:full-scores} and \ref{table:confusion-matrix} detail the results of classification for each label in terms of performance scores (precision, recall and F1) and confusion matrix, respectively (for our HSLN-RNN model trained on the PubMed 20k dataset). These show that the classifier is very good at predicting the labels \textit{Methods}, \textit{Results} and \textit{Conclusions}, whereas the greatest difficulty the classifier has is in distinguishing \textit{Background} sections from \textit{Objectives} sections. One fifth of \textit{Background} sentences are incorrectly classified as \textit{Objectives}, while around one forth of \textit{Objectives} sentences are wrongly assigned to the label of \textit{Background}. We conjecture this difficulty mainly comes from the fact that the difference between \textit{Background} and \textit{Objectives} sentences in terms of writing style is less obvious compared with the other sections of the abstract. Moreover, our model has some difficulty in telling \textit{Methods} sentences apart from \textit{Results} sentences.
%
%
%
%
%
%

\begin{table}[h!]
\centering
\begin{tabulary}{0.5\textwidth}{lccccp{0em}}
\hlineB{2.5}
\textbf{Label} & P & R & F1 & Support \\ \hline
Background         & 78.5               & 80.0            & 79.2        & 3077          \\ 
Objectives      & 74.2               & 69.9            & 72.0        & 2333             \\ 
Methods     & 95.0               & 97.7            & 96.3        & 9884             \\ 
Results           & 96.8               & 95.3            & 96.0        & 9713             \\  
Conclusions        & 97.6               & 96.5            & 97.1        & 4571             \\ \hline
Total             & 92.6               & 92.7            & 92.6        & 29578            \\ \hlineB{2.5}
\end{tabulary}
\caption{Results (presented in percentage) in terms of precision (P), recall (R) and F-measure (F1) on the test set for each label obtained by our HSLN-RNN model on the PubMed 20k dataset.}
\label{table:full-scores}
\end{table}

\begin{table}[ht!]
\centering
\resizebox{\columnwidth}{!}{%
\begin{tabular}{|l|c|c|c|c|c|}
\hline
\textbf{}  & \textbf{B} & \textbf{C} & \textbf{M} & \textbf{O} & \textbf{R}  \\ \hline
\textbf{B} & 2460       & 4          & 69          & 537        & 7                 \\ \hline
\textbf{C} & 4          & 4413       & 11          & 1          & 142                 \\ \hline
\textbf{M} & 37         & 11         & 9657       & 27          & 152                  \\\hline
\textbf{O} & 632        & 0         & 68          & 1630       & 3                    \\ \hline
\textbf{R} & 2          & 95         & 362        & 1          & 9253               \\ \hline
\end{tabular}
}
\caption{Confusion matrix obtained by our model on the PubMed 20k dataset. Rows correspond to predicted labels, and columns correspond to true labels. B represents background, O represents objectives, M represents methods, R represents results, and C represents conclusions.}
\label{table:confusion-matrix}
\end{table}

Table \ref{table:error-analysis} presents a few examples of prediction errors that are produced by our HSLN-RNN model trained on the PubMed 20k dataset. This error analysis suggests that one of the biggest model error sources could be from the debatable gold standard labels of the dataset. For example, the sentence ``Depressive disorders are one of the leading components of the global burden of disease with a prevalence of up to 14\% in the general population.'' is indeed introducing the background of the problem (depressive disorders) on which this article is going to focus; however, the gold label classifies it into the \textit{Objective} category. For another instance, the sentence ``A post hoc analysis was conducted with the use of data from the evaluation study of congestive heart failure and pulmonary artery catheterization effectiveness (escape).'' belongs to the \textit{Result} label according to the gold standard, but it makes more sense that it should be classified as a \textit{Method} label.

\begin{table*}[t]
\centering
\begin{tabular}{ |>{\raggedright\arraybackslash} m{11.2cm} |>{\centering\arraybackslash} m{1.75cm} |>{\centering\arraybackslash} m{1.75cm}|}
\hline
\textbf{Sentence}                                     & \textbf{Predicted} & \textbf{Gold} \\ \hline

Depressive disorders are one of the leading components of the global burden of disease with a prevalence of up to 14\% in the general population. [25829103]                     & Background                  & Objective             \\ \hline
This study assessed whether diets with different fat quality and supplementation with coenzyme Q10 (CoQ) affect the metabolomic profile in urine. [24986061]                                                 & Objective                  & Background             \\ \hline
A post hoc analysis was conducted with the use of data from the evaluation study of congestive heart failure. [24845963]                                                                                                                         & Method                  & Result             \\ \hline
Hence, 47 secondary schools from all 12 districts of the city [...] are participating in the study. [25150368] & Result                  & Method             \\ \hline
This study investigated whether oxytocin can affect attentional bias in social anxiety. [25552432]                                                                                                                   & Objective                  & Method             \\ \hline
We hypothesize that BMC + Phone and BMC + Home will produce greater reductions in BMI percentiles than BMC alone. [24456698]                                                                                                                                                  & Conclusion                  & Method             \\ \hline
\end{tabular}
\caption{Examples of prediction errors of our HSLN-RNN model trained on the PubMed 20k dataset. Each sentence is followed by the PMID of the abstract that this sentence belongs to, which is enclosed in middle brackets. The ``Predicted'' column indicates the label predicted by our model for a given sentence. The ``Gold'' column indicates the gold label of the sentence.}
\label{table:error-analysis}
\end{table*}

\begin{figure}[h!]
\begin{center}
\includegraphics[width=0.5\textwidth]{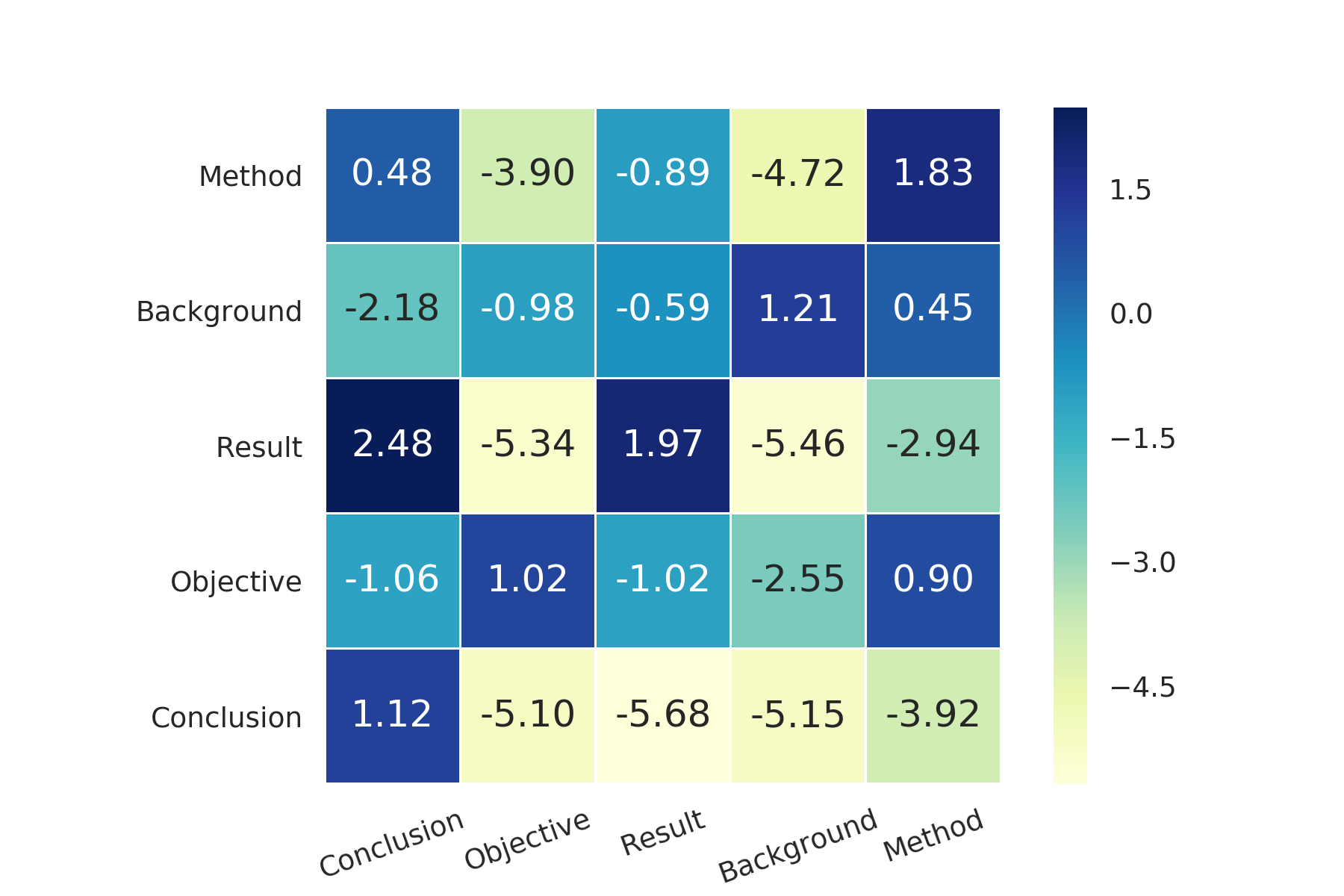}
\caption{Transition matrix of label sequence after the HSLN-RNN model has been trained on the PubMed 20k dataset. The rows represent the label of the previous sentence, while the columns represent the label of the current sentence.}
\label{figure:heatmap}
\end{center}
\end{figure}

Figure \ref{figure:heatmap} presents an example of the transition matrix after the HSLN-RNN model has been trained on the PubMed 20k dataset, which encodes the transition probability between two subsequent labels. It effectively reflects what label is the most likely one that follows the current one. For example, by comparing the transition scores in the \textit{Result} row in Figure \ref{figure:heatmap}, we can conclude that a sentence pertaining to the \textit{Result} is typically followed by a sentence pertaining to the \textit{Conclusion} and is unlikely to be followed by a sentence in the \textit{Background} category (transition scores of 2.48 vs -5.46), which makes sense. From this transition matrix, we can figure out the most probable label sequence: $Background\to Objective\to Method\to Result\to Conclusion$, which is also consistent with our expectations.
%
%

In order to test the importance of pretrained word embeddings, we performed experiments with different sets of publicly published word embeddings, as well as our locally curated word embeddings, to initialize our model. Table \ref{table:embeddings} gives the performance of six different word embeddings for our HSLN-RNN model trained on the PubMed 20k dataset. According to Table \ref{table:embeddings}, the training methods that create the word embeddings do not have a strong influence on model performance, but the corpus they are trained on does. The combination of Wikipedia and PubMed abstracts as the corpus for unsupervised word embedding training yields the best result, and the individual use of either the Wikipedia corpus or the PubMed abstracts performs much worse. Although the dataset we are using for evaluation is also from PubMed abstracts, using only the PubMed abstracts together with MIMIC notes without the Wikipedia corpus does not guarantee better result (see the ``FastText-P.M.+MIMIC'' embeddings in Table \ref{table:embeddings}), which may be because the corpus size of PubMed abstracts plus MIMIC notes (about 12.8 million abstracts and 1 million notes) is not large enough for good embedding training compared with the corpus consisting of at least billion tokens such as the Wikipedia. 
%
%

\begin{table}[h!]
\centering
\begin{tabular}{llr}
\hlineB{2.5}
Embedding            & Dimension & P.M. 20k \\ \hline
Glove-wiki                & 200       & 92.0       \\
FastText-wiki        & 300       & 92.2       \\
FastText-P.M.+MIMIC  & 300       & 92.0       \\
Word2vec-News        & 300       & 92.2       \\
Word2vec-wiki        & 200       & 92.1       \\
Word2vec-wiki+P.M.   & 200       & \textbf{92.6}       \\\hlineB{2.5}
\end{tabular}
\caption{Comparison of performance with different choices of word embeddings for our HSLN-RNN model trained on the PubMed 20k dataset (reported on F1-scores on the test set). ``P.M.'' means PubMed.}
\label{table:embeddings}
\end{table}

\section{Conclusion}

In this work, we have presented an ANN based hierarchical sequential labeling network to classify sentences that appear sequentially in text. We demonstrate that incorporating the contextual information from surrounding sentences to help classify the current one by using an LSTM layer to sequentially process the encoded sentence representations can improve the overall quality of predictions. Our model outperforms the state-of-the-art results by 2\%-3\% on two datasets for sequential sentence classification in medical abstracts. We expect that our proposed model can be generalized to any problem that is related to sequential sentence classification, such as the paragraph-level sequential sentence categorization in full-text articles for better text mining and document retrieval \cite{westergaard2018comprehensive}.

\section{Future Work}

Although the whole PubMed database contains over 2 million abstracts with part of them accompanied by full-text articles, only a small fraction of them are structured and contain the label information utilized in this work. We plan to make use of the rest unannotated abstracts or full texts to pre-train our model and then fine tune it to the target annotated datasets inspired by the work from \cite{howard2018universal} so that the performance can be further boosted. 

\section*{Acknowledgments}

This work was supported by funding grant U54-HG007963 from National Human Genome Research Institute (NHGRI).




\bibliography{emnlp2018}
\bibliographystyle{acl_natbib_nourl}


\end{document}